\documentclass[11pt,a4paper]{article}
\usepackage[hyperref]{acl2020}
\usepackage{times}
\usepackage{latexsym}

\usepackage{microtype}

\usepackage{graphicx} 
\usepackage{multirow} 
\usepackage{booktabs} 
\usepackage{amsmath}
\usepackage{enumitem}
\usepackage{arydshln}
\usepackage{subcaption}
\usepackage{amssymb}
\usepackage{kotex}
\usepackage{tablefootnote}

\newcommand\Tstrut{\rule{0pt}{2.5ex}}         

\aclfinalcopy 
\def\aclpaperid{1079} 

\title{Fast and Accurate Deep Bidirectional Language Representations \\for Unsupervised Learning}

\author{Joongbo Shin, Yoonhyung Lee, Seunghyun Yoon, Kyomin Jung \\
  Seoul National University\\
  Republic of Korea \\
  {\texttt \{jbshin,~cpi1234,~mysmilish,~kjung\}@snu.ac.kr}\\}

\date{}

\begin{document}
\maketitle

\begin{abstract}

Even though BERT achieves successful performance improvements in various supervised learning tasks, applying BERT for unsupervised tasks still holds a limitation that it requires repetitive inference for computing contextual language representations.
To resolve the limitation, we propose a novel deep bidirectional language model called \textbf{T}ransformer-based \textbf{T}ext \textbf{A}utoencoder (T-TA).
The T-TA computes contextual language representations without repetition and has benefits of the deep bidirectional architecture like BERT.
In run-time experiments on CPU environments, the proposed T-TA performs over six times faster than the BERT-based model in the reranking task and twelve times faster in the semantic similarity task.
Furthermore, the T-TA shows competitive or even better accuracies than those of BERT on the above tasks\footnote{Code is available at https://github.com/joongbo/tta}.

\end{abstract}

\section{Introduction}

A language model is an essential component in many NLP applications ranging from automatic speech recognition (ASR)  \cite{chan2016listen,panayotov2015librispeech} to neural machine translation (NMT) \cite{sutskever2014sequence,sennrich2016improving,vaswani2017attention}.
Recently, BERT (Bidirectional Encoder Representations from Transformers) \cite{devlin2019bert} and its variations have brought significant improvements in learning natural language representation, and they have achieved state-of-the-art performances on various downstream tasks such as GLUE benchmark \cite{wang2019glue} and question answering \cite{rajpurkar2016squad}.
This success of BERT continues in various unsupervised tasks such as the $N$-best list reranking for ASR and NMT \cite{shin2019effective,salazar2019pseudolikelihood}, showing that deep bidirectional language models are useful in unsupervised applications as well.

However, when applying the BERT to unsupervised learning tasks, there exists significant inefficiency in computing language representations at the inference stage \cite{salazar2019pseudolikelihood}.
During training, the BERT uses the \textit{masked language modeling} (MLM) objective, which is to predict the original ids of explicitly masked words from the input.
Due to the MLM objective, each contextual word representation should be computed by a two-step process of masking a word in the input and feeding it into the BERT.
During inference, therefore, this process repeats $n$ times to obtain obtain the representations of the whole words of a text sequence  \cite{wang2019bert,shin2019effective,salazar2019pseudolikelihood}, resulting in computational complexity of $O(n^3)$\footnote{$O(n^2)$ is from the per-layer complexity of Transformer \cite{vaswani2017attention}.} in terms of the number $n$ of words.
Hence, it is necessary to reduce the computational complexity when we apply the model to the case where the inference time is considered critical, \textit{e.g.} mobile environments and real-time systems~\cite{sanh2019distilbert,lan2019albert}.
Faced with this limitation of the BERT, we raise a new research question: 
``Can we make a deep bidirectional language model that has minimal inference time while maintaining the accuracy of BERT?''

In this paper, we answer "YES" to the above question by proposing a novel bidirectional language model named \textbf{T-TA}: \textbf{T}ransformer-based \textbf{T}ext \textbf{A}utoencoder that has the reduced computational complexity of $O(n^2)$ when applying the model to the unsupervised applications.
The proposed model is trained with a new learning objective named \textit{language autoencoding} (LAE). 
The LAE let the target labels to be the same as the text input, and its objective is to predict every token in the input sequence at once without merely copying the input to the output.
To learn the proposed objective, we devise a \textbf{diagonal masking} operation and an \textbf{input isolation} mechanism inside the T-TA based on the Transformer encoder \cite{vaswani2017attention}.
These components enable the proposed T-TA to compute contextualized language representations at once while maintaining the benefits of the deep bidirectional architecture of BERT.

We conduct a series of experiments on two unsupervised tasks: the $N$-best list reranking and the unsupervised semantic textual similarity.
First, in the runtime experiments on CPU environments, we show that the proposed T-TA is $6.35$ times faster than the BERT-based model in the reranking task, and $12.7$ times faster in the semantic similarity task.
Second, even with this faster inference, the T-TA achieves competitive performances to BERT on reranking tasks.
Furthermore, the T-TA outperforms BERT up to 8 points in Pearson's $r$ on unsupervised semantic textual similarity tasks.

\section{Related Works}

When referring to the autoencoder for language modeling, sequence-to-sequence learning approaches have been commonly used.
These approaches encode a given sentence into a compressed vector representation, followed by a decoder which reconstructs the original sentence from the \textit{sentence-level} representation \cite{sutskever2014sequence,cho2014learning,dai2015semi}.
To the best of our knowledge, however, none of them considered an autoencoder that encodes \textit{word-level} representations like BERT without the autoregressive decoding process.

There have been many studies on neural network-based language models for word-level representations.
Distributed word representations were proposed and gained huge interests as they were considered to be fundamental building blocks for the natural language processing tasks~\cite{rumelhart1986learning,bengio2003neural,mikolov2013distributed}. 
Recently, researchers explored contextualized representations of text where each word will have different representations depending on the context~\cite{peters2018deep,radford2018improving}.
More recently, the Transformer-based deep bidirectional model was proposed and applied to the various supervised-learning tasks with a huge success~\cite{devlin2019bert}.

For unsupervised tasks, researchers adopted the recent language-representation models and investigated their effectiveness. 
One typical example is the $N$-best list reranking for ASR and NMT tasks.
In particular, there have been researches integrating the left-to-right and the right-to-left language models \cite{arisoy2015bidirectional,chen2017investigating,peris2015bidirectional} so as to outperform conventional unidirectional language models \cite{mikolov2010recurrent,sundermeyer2012lstm} in these tasks.
Furthermore, BERT-based approaches have been explored and have achieved significant performance improvements on these tasks based on the fact that; bidirectional language models yield the pseudo-log-likelihood of a given sentence; this score is useful in ranking the $n$-best hypotheses~ \cite{wang2019bert,shin2019effective,salazar2019pseudolikelihood}.

Another line of research includes reducing the computation time and memory consumption of BERT. 
\citet{lan2019albert} proposed parameter-reduction techniques, factorized embedding parameterization and cross-layer parameter sharing, and achieved 18 times fewer parameters and $1.7$ times faster training time. With a similar research direction, \citet{sanh2019distilbert} presented a method to pre-train a smaller model that can be finetuned for the downstream task, and achieved a $1.4$ times lower parameter count with $1.6$ times faster inference. However, none of these studies presented methods that directly revise BERT architecture for decreasing computational complexity during inference.

\section{Language Model Baselines}
\begin{figure*}[t]
    \centering
    \begin{subfigure}{.3\textwidth}
        \centering
        \includegraphics[width=.8\linewidth]{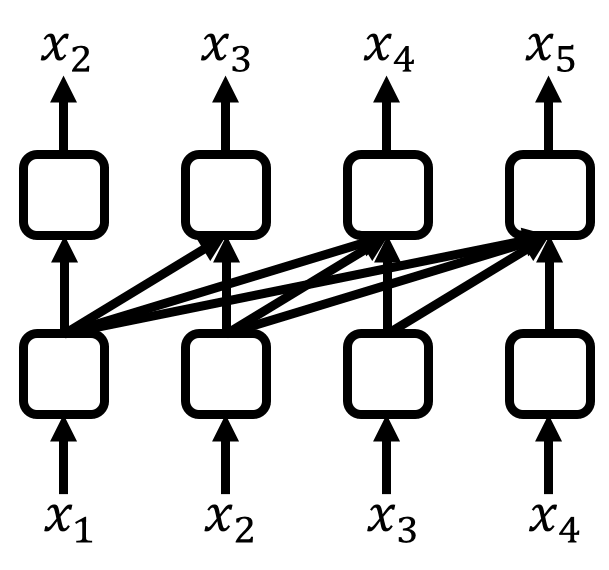}
        \caption{Causal language modeling}
        \label{fig:clm}
    \end{subfigure}
    \begin{subfigure}{.3\textwidth}
        \centering
        \includegraphics[width=.8\linewidth]{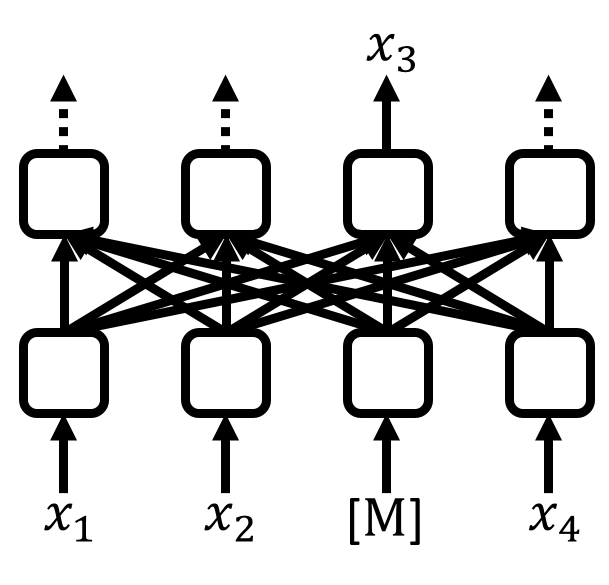}
        \caption{Masked language modeling}
        \label{fig:mlm}
    \end{subfigure}
    \begin{subfigure}{.3\textwidth}
        \centering
        \includegraphics[width=.8\linewidth]{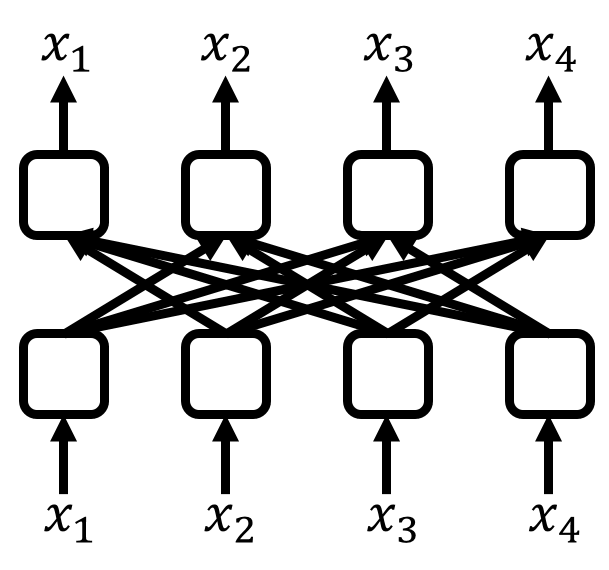}
        \caption{Language Autoencoding}
        \label{fig:lae}
    \end{subfigure}
    \caption{Schematic diagrams of Transformer-based language models for (a) CLM, (b) MLM, and (c) LAE.}
    \label{fig:lms}
\end{figure*}
The conventional language modeling is a task of predicting the $i$-th token $x_i$ using its preceding context $\textbf{x}_{<i}\,{=}\,[x_1,\ldots,x_{i-1}]$, and we call this objective as causal language modeling (CLM) throughout this paper following \cite{conneau2019cross}.
As shown in Figure \ref{fig:clm}, we can obtain (left-to-right) contextualized language representations $\textbf{H}^\text{C}\,{=}\,[H^\text{C}_1,\ldots,H^\text{C}_n]$ at a single feeding the input sequence to the CLM-trained language model, where $H^\text{C}_i\,{=}\,h^\text{C}(\textbf{x}_{<i})$ is the hidden representation of $i$-th token.
This paper takes this unidirectional language model (uniLM) as our speed baseline.
However, contextualized language representations obtained from the uniLM are insufficient to accurately encode a given text because future contexts cannot be used to understand the current tokens during inference.

Recently, BERT \cite{devlin2019bert} enables the full contextualization of the language representations by using the masked language modeling (MLM) objective.
In the MLM, some tokens from the input sequence are randomly masked, and the objective is to predict the original tokens at the masked positions using only their context.
As in Figure \ref{fig:mlm}, we can obtain a contextualized representation of $i$-th token $H^\text{M}_i\,{=}\,h^\text{M}(M_i({\textbf{x}}))$ by masking the token in the input sequence and feeding it into the MLM-trained model, where $M_i({\textbf{x}})\,{=}\,[x_1,\ldots,x_{i-1},\texttt{[MASK]},x_{i+1},\ldots,x_n]$ is an external masking operation.
This paper takes this bidirectional language model (biLM) as our performance baseline.
However, this \textit{mask-and-predict} approach should be repeated $n$ times to obtain the whole language representations because the learning occurs only at the masked position during the MLM training.
Although the language representations are robust and accurate, this repetition causes significant inefficiency in the use of unsupervised applications such as the $N$-best list reranking tasks \cite{wang2019bert,shin2019effective,salazar2019pseudolikelihood}.

\section{Proposed Methods}

\subsection{Language Autoencoding}

In this paper, we propose a new learning objective named \textit{language autoencoding} (LAE) for obtaining fully contextualized language representations without repetition.
The LAE lets the output to become the same as the input, and the objective is to predict every token in a text sequence at once without merely copying the input to output.
For the proposed task, a language model should reproduce the whole input at once while avoiding the over-fitting. Otherwise, the model only outputs the representation copied from the input representation without learning any statistics of the language.
To this end, information flow from the $i$-th input to the $i$-th output should be blocked inside the model shown in Figure \ref{fig:lae}.
From this LAE objective, we can obtain fully contextualized language representations $\textbf{H}^\text{L}\,{=}\,[\textbf{H}^\text{L}_1,\ldots,\textbf{H}^\text{L}_n]$ at once, where $\textbf{H}^\text{L}_i\,{=}\,\textbf{H}^\text{L}(\textbf{x}_{\text{\textbackslash}i})$ and $\textbf{x}_{\text{\textbackslash}i}\,{=}\,[x_1,\ldots,x_{i-1},x_{i+1},\ldots,x_n]$.
The way of blocking the information flow is described in the next section.

\subsection{Transformer-based Text Autoencoder}
\label{ssec:t-ta}

\begin{figure}[t]
    \centering
    \includegraphics[width=.65\linewidth]{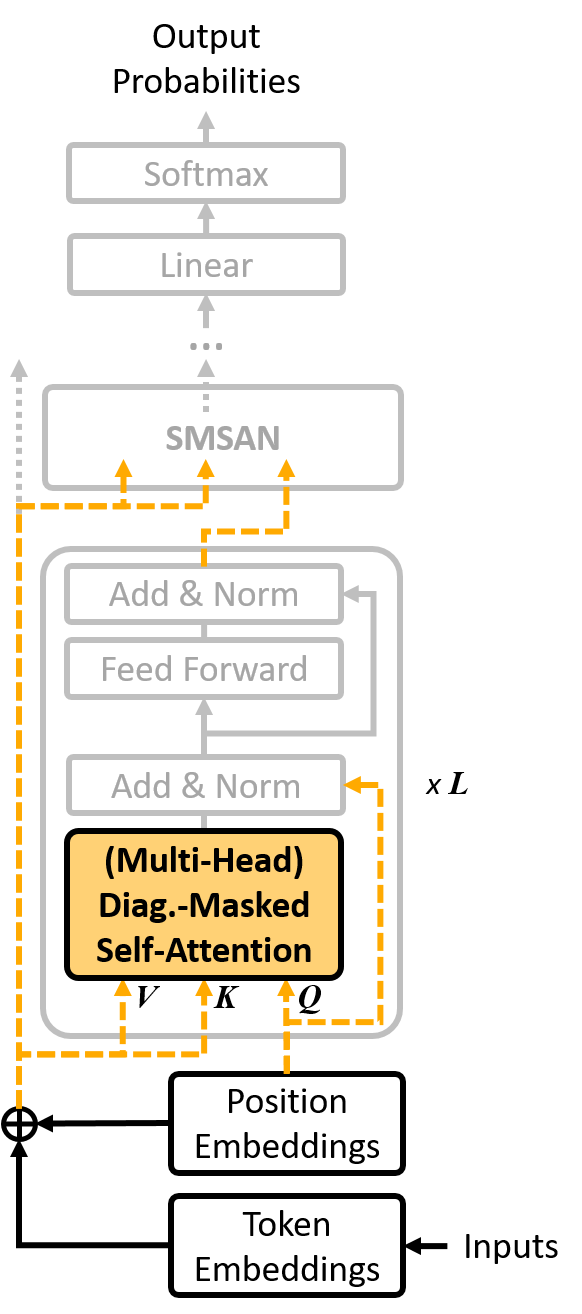}
    \caption{Architecture of our T-TA. Highlighted box and dashed arrows are newly invented in this paper.}
    \label{fig:arch}
\end{figure}

In this section, we introduce a novel architecture of a deep bidirectional language model named \textbf{T-TA}, which stands for \textbf{T}ransformer-based \textbf{T}ext \textbf{A}utoencoder, and the overall architecture of the T-TA is shown in Figure \ref{fig:arch}.
As in its name, the model architecture is based on the Transformer encoder \cite{vaswani2017attention}.
To learn the proposed LAE, we develop a \textbf{diagonal masking} operation and an \textbf{input isolation} mechanism inside the T-TA.
Both developments are designed to let the language model predict every token at once while maintaining the deep bidirectional property (see the descriptions in the following subsections).
Due to the space limit, we refer to the original Transformer paper \cite{vaswani2017attention} for other details of the standard functions such as the multi-head attention, the scaled dot-product attention, layer normalization, and the position-wise fully connected feed-forward network.

\subsubsection{Diagonal Masking}
\label{sssec:diagonal}
\begin{figure}[t]
    \centering
    \includegraphics[width=.65\linewidth]{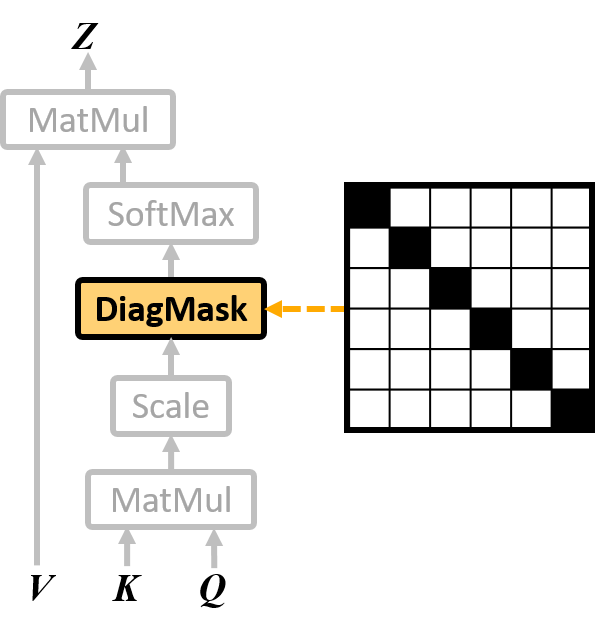}
    \caption{Diagonal masking of the scaled dot-product attention mechanism. Highlighted box and dashed arrow are newly invented in this paper.}
    \label{fig:dmsa}
\end{figure}

As shown in Figure \ref{fig:dmsa}, a diagonal masking operation is inside the scaled dot-product attention in order to be ``self-unknown'' during the inference.
This operation prevents the information from flowing to the same position in the next layer by masking out the diagonal values in the input of the softmax.
Specifically, the output vector at each position is the weighted sum of the value \textbf{V} at other positions, where the attention weights come from the query \textbf{Q} and the key \textbf{K}.

The diagonal mask becomes meaningless when we use it together with the residual connection or utilize it in the multi-layer architecture.
To keep the self-unknown functional, we can remove the residual connection and adopt single-layer architecture. 
However, it is essential to utilize deep architecture to understand the intricate patterns of natural language.
To this end, we further develop an architecture described in the next section.

\subsubsection{Input Isolation}
\label{sssec:isolation}

We now propose an input isolation mechanism in order to make the residual connection and the multi-layer architecture compatible with the diagonal masking operation.
In the input isolation, the key-value inputs (\textbf{K}-\textbf{V}) of all encoding layers are isolated from the network flow, and they are fixed to the sum of the token embeddings and the position embeddings.
Only query inputs (\textbf{Q}) are updated across the layers during inference by referring to the fixed output of the embedding layer.

Additionally, we input the position embeddings to the Q of the very first encoding layer in order to make the self-attention mechanism effective.
Otherwise, the attention weights will be the same at all positions, resulting in that the first self-attention works as a simple calculator of averaging the input representations except the ``self'' position.
Finally, we utilize the residual connection only to the query to maintain the unawareness completely.
The dashed arrows in Figure~\ref{fig:arch} show this input isolation mechanism inside the T-TA.

By using the diagonal masking and input isolation together, the T-TA can have multiple encoder layers.
They enable the T-TA to obtain high-quality contextual language representations with an only single feeding of a sequence.

\subsection{Discussion and Analysis}

Until now, we have introduced the new learning objective, language autoencoding (LAE), and the novel deep bidirectional language model, Transformer-based Text Autoencoder (T-TA).
We will verify the model architecture of the proposed T-TA in Section \ref{sssec:verify}, and compare our model with the recent bidirectional language model BERT in Section \ref{sssec:comp-bert}.

\subsubsection{Verification of the Architecture}
\label{sssec:verify}

We here discuss how the diagonal masking with input isolation preserve ``self-unknown'' property in detail.

As in Figure \ref{fig:arch}, we have two embeddings, token embeddings $\textbf{X}\,{=}\,[X_1,\ldots,X_n]^T\in \mathbb{R}^{n\times d}$ and position embeddings $\textbf{P}\,{=}\,[P_1,\ldots,P_n]^T\in \mathbb{R}^{n\times d}$, where the $d$ is an embedding dimension.
From the input isolation, the key and value $\textbf{K}\,{=}\,\textbf{V}\,{=}\,\textbf{X}+\textbf{P}$ have the information of input tokens and they are \textit{fixed} in all layers, but the query $\textbf{Q}^{l}$ is \textit{updated} across the layers during inference started from the position embeddings $\textbf{Q}^{1}\,{=}\,\textbf{P}$ at the first layer.

Let us consider the $l$-th encoding layer's query input $\textbf{Q}^{l}$ and its output $\textbf{H}^{l}\,{=}\,\textbf{Q}^{l+1}$. Then,
\begin{equation}
\begin{split}
    \textbf{H}^{l} &= \text{SMSAN}(\textbf{Q}^{l},\textbf{K},\textbf{V})\\
    &= g(\text{Norm}(\text{Add}(\textbf{Q}^{l},f(\textbf{Q}^{l},\textbf{K},\textbf{V})))),
\end{split}
\end{equation}
where $\text{SMSAN}(\cdot)$ represents the Self-Masked Self-Attention Network, the encoding layer of the T-TA, $g(x)\,{=}\,\text{Norm}(\text{Add}(x,\text{FeedForward}(x)))$, two upper-side sub-boxes of the encoding layer, and $f(\cdot)$ is the (multi-head) diagonal-masked self-attention (DMSA) mechanism shown in Figure \ref{fig:arch}.
As in Figure \ref{fig:dmsa}, the DMSA module computes $\textbf{Z}^l$ as follows:
\begin{equation}
\begin{split}
    \textbf{Z}^l &= f(\textbf{Q}^{l},\textbf{K},\textbf{V}) = \text{DMSA}(\textbf{Q}^{l},\textbf{K},\textbf{V})\\
    &= \text{SoftMax}(\text{DiagMask}({\textbf{Q}^{l}\textbf{K}^{T}}/{\sqrt{d}}))\textbf{V}.\\
\end{split}
\end{equation}

In the DMSA module, the $i$-th element of $\textbf{Z}^l\,{=}\,[Z^l_1,\ldots,Z^l_n]^T$ is always computed by a weighted average of the fixed $\textbf{V}$ discarding the information of $i$-th token $X_i$ in $V_i$.
To be more specific, $Z^l_i$ is the weighted average of the $\textbf{V}$ with the attention weight vector $\textbf{s}^l_i$, \textit{i.e.}, $Z^l_i\,{=}\,\textbf{s}^l_i\textbf{V}$, where $\textbf{s}^l_i\,{=}\,[s^l_1,\ldots,s^l_{i-1},0,s^l_{i+1},\ldots,s^l_n]\,{\in}\,\mathbb{R}^{1\times n}$.
We here note that only the DMSA is related to the ``self-unknown" since no token representation is referred to each other in the subsequent transformations from $\textbf{Z}^l$ to $\textbf{H}^l$.
Therefore, it is guaranteed that the $i$-th element of the query representation in any layer, $Q^l_i$, never sees the corresponding token representation started from the ${Q}^1_{i}\,{=}\,P_i$.
Consequently, the T-TA preserves the ``self-unknown'' property during inference while maintaining the residual connection and multi-layer architecture.


\subsubsection{Comparison with BERT}
\label{sssec:comp-bert}

There are several differences between the strong baseline BERT \cite{devlin2019bert} and the proposed model T-TA, while both models learn deep bidirectional language representations.
\begin{itemize}[leftmargin=*]
    \item
    While BERT uses external masking operation in the input, T-TA has internal masking operation in the model as we intend.
    Also, while BERT is based on denoising autoencoder, T-TA is based on autoencoder.
    Due to this novel approach, the T-TA does not need \textit{mask-and-predict} repetition during computing contextual language representations.
    Consequently, we reduce the computational complexity from $O(n^3)$ of BERT to $O(n^2)$ of T-TA when applying the language models to the unsupervised learning tasks.
    
    \item
    As in the T-TA, feeding an intact input (without masks) into BERT is also possible.
    However, we argue that it will significantly hurt the model performance on unsupervised applications since the MLM objective does not consider the intact token much.
    We include experiments that show model performance with the intact input (described in Table \ref{tab:wers}, \ref{tab:sts}, and \ref{tab:sick}).
    We also suggest reading previous research that reported the same opinion~\cite{salazar2019pseudolikelihood}.
    
\end{itemize}

\section{Experiments}

To evaluate the proposed method, we conduct a series of experiments. We first evaluate the contextual language representations obtained from the Transformer-based Text Autoencoder (T-TA) on the $N$-best list reranking tasks. We then apply our method to unsupervised semantic textual similarity (STS) tasks. The following sections will demonstrate that the proposed model is much faster than the BERT during inference (in Section \ref{ssec:runtime}) while showing competitive or even better accuracies than those of the BERT on reranking tasks (in Section \ref{ssec:reranking}) and STS tasks (in Section \ref{ssec:sts}).

\subsection{Language Model Setups}
\label{ssec:setup}

This paper mainly compares the proposed T-TA with the bidirectional language model (biLM), which is trained with the masked language modeling (MLM) objective, like BERT.
For a fair comparison, each model has the same number of parameters based on the Transformer as followed: $|L|=3$ self-attention layers with $d=512$ input and output dimensions, $h=8$ attention heads, and $d_f=2048$ hidden units for the position-wise feed-forward layers.
We use a \textit{gelu} activation \cite{hendrycks2016bridging} rather than the standard \textit{relu}, following OpenAI GPT \cite{radford2018improving} and BERT \cite{devlin2019bert}.
We set a position embeddings to be trainable following BERT \cite{devlin2019bert} rather than a fixed sinusoid \cite{vaswani2017attention} with supported sequence lengths up to 128 tokens in our experiments.
We use WordPiece embeddings \cite{wu2016google} with a vocabulary of about $|V|\simeq30,000$ tokens.
The weights of the embedding layer and the last softmax layer of the Transformer are shared.
For the speed baseline, we also implement a unidirectional language model (uniLM), which has the same number of parameters as T-TA and biLM.

For training, we make a training instance consisting of a single sentence with \texttt{[BOS]} and \texttt{[EOS]} tokens at the begin and the end of each sentence.
We use 64 sentences as a training batch, and train language models $1M$ steps for ASR and $2M$ steps for NMT.
We train the language models with Adam \cite{kingma2014adam} with an initial learning rate of $1e-4$, $\beta_1=0.9$, $\beta_2=0.999$, learning rate warm up over the first $50$k steps and linear decay of the learning rate.
We use a dropout probability of 0.1 on all layers.
Our implementation is based on Google's official code for BERT\footnote{https://github.com/google-research/bert}.

To train language models that we implement, we use about 13GB English Wikipedia dump that has about 120M sentences.
The trained models are used for reranking in neural machine translation (NMT) and unsupervised semantic textual similarity tasks.
For reranking in automatic speech recognition (ASR), we use additional in-domain training data of the 4.0GB normalized text data of the official LibriSpeech corpus that has about 40M sentences.

One of the strong baseline language models, the pre-trained BERT-base-uncased \cite{devlin2019bert}, is used for reranking and STS.
We also include the reranking results from the traditional count-based $5$-gram language models that are trained on each dataset using the KenLM library \cite{heafield2011kenlm}.

\subsection{Running Time Analysis}
\label{ssec:runtime}

\begin{figure}[t]
    \centering
    \includegraphics[width=\linewidth]{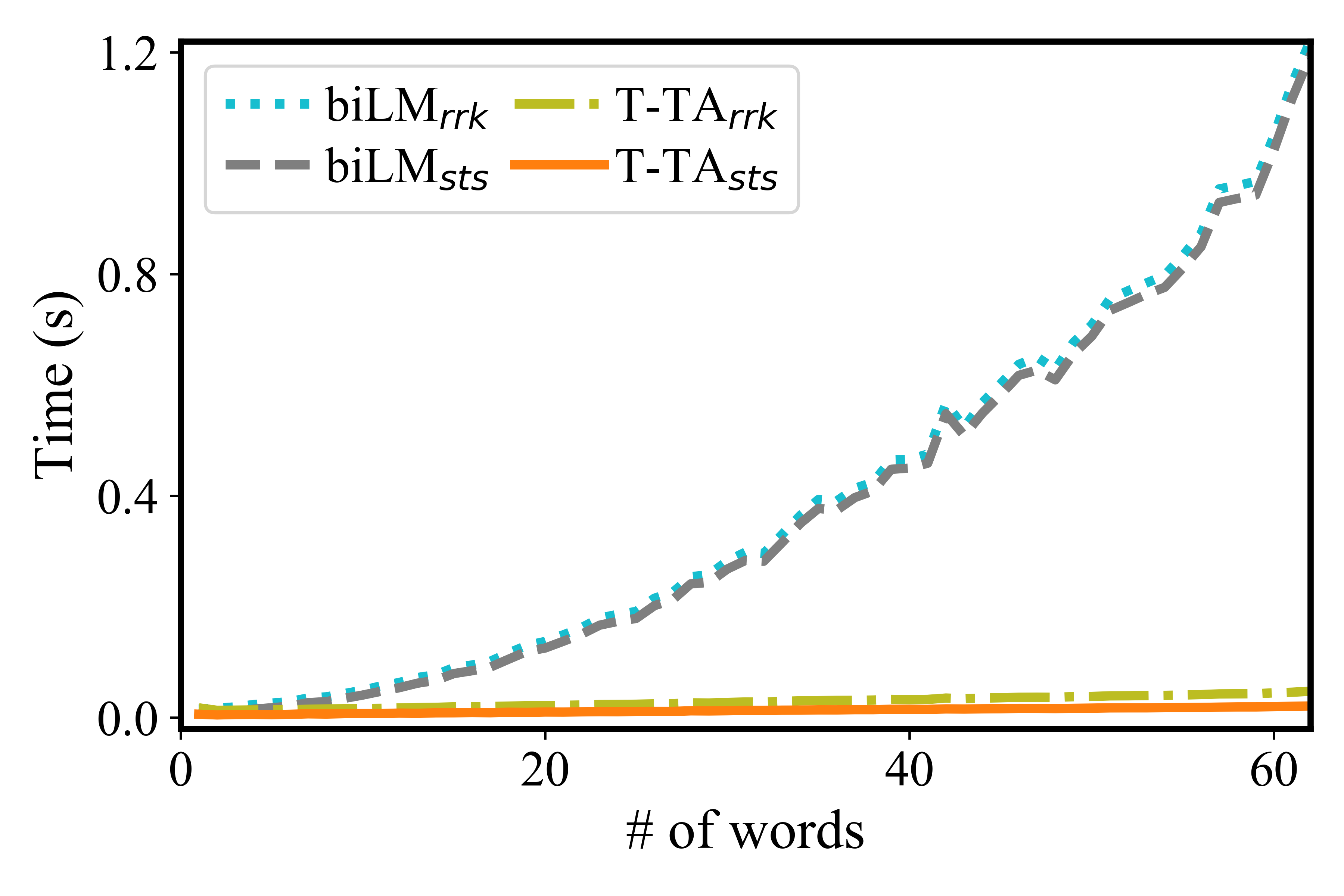}
    \caption{Average running times of each model according to the number of words on STS and reranking tasks, sub-scripted as \textit{sts} and \textit{rrk} respectively.}
    \label{fig:times}
\end{figure}

We first measure the running time of each language model for computing the contextual language representation $\textbf{H}^{L}\,{\in}\, \mathbb{R}^{n\times d}$ of a given text sequence.
In the unsupervised STS tasks, we directly use the $\textbf{H}^{L}$ for the analysis. 
In the case of the reranking task, we need further computation; we compute $\text{Softmax}(\textbf{H}^{L} \textbf{E}^T)$ to obtain the likelihood of each token, where $\textbf{E}\in \mathbb{R}^{|V|\times d}$ is the weight parameters of the softmax layer.
Therefore, the computational complexity of reranking is bigger than that of STS.

In the running time measurement, we use Intel(R) Core(TM) i7-6850K CPU (3.60GHz) on the TensorFlow 1.12.0 library with Python 3.6.8 over the Ubuntu 16.04.06 LTS.
In each experiment, we measured the run-time 50 times and averaged the results.
Figure \ref{fig:times} shows that the run-time of the T-TA is faster than that of the biLM, and it becomes significant as the sentence is longer.
For numerical comparison, we set the standard number of words to $20$ since the average number of words in English sentences today is about $20$ \cite{dubay2007classic}.
In this setup, the T-TA takes about $9.85$ ms while the biLM takes about $125$ ms in the STS task, showing that the T-TA is $12.7$ times faster than the biLM.
In the reranking task, the ratio between the T-TA and the biLM is reduced to $6.35$ times (still significant), and this is because the repetition of biLM is related only to computing $\textbf{H}^{L}$ not to $\text{Softmax}(\textbf{H}^{L} \textbf{E}^T)$.

For the visibility of Figure \ref{fig:times}, we omit the run-time results of uniLM, which is also as fast as the T-TA (Appendix \ref{a:ssec:runtime}).
With this fast inference, we show that the T-TA is as accurate as BERT in the next section.

\subsection{Reranking the N-best List}
\label{ssec:reranking}
To evaluate language models, we conduct experiments on the unsupervised task of reranking the $N$-best list.
In the experiments, we apply each language model to rerank the $50$-best candidate sentences, which are obtained in advance using each sequence-to-sequence model on ASR and NMT.
The ASR and NMT models we implement are detailed in Appendix \ref{a:ssec:asr} and \ref{a:ssec:nmt}.

We rescore the sentences by linearly interpolating two scores from a sequence-to-sequence model and each language model as follows:
\begin{equation*}
\label{eq:inter2}
    \text{score}=(1-\lambda)\cdot\text{score}_{s2s}+\lambda\cdot\text{score}_{lm},
\end{equation*}
where the $\text{score}_{s2s}$ is the score from sequence-to-sequence models, $\text{score}_{lm}$ is the score from language models calculated by the sum (or mean) of the log-likelihood of each token, and the interpolation weight $\lambda$ is set to a value that shows the best performance in the development set.

We note that the T-TA and biLM (also BERT) assign the pseudo-log-likelihood to the score of a given sentence while the uniLM assigns the log-likelihood.
Because the reranking task is based on relative scores of the $n$-best hypotheses, the fact that bidirectional language models yield the pseudo-log-likelihood of a given sentence does not matter in this task \cite{wang2019bert,shin2019effective,salazar2019pseudolikelihood}.

\subsubsection{Results on Speech Recognition}
For reranking in ASR, we use prepared $N$-best lists obtained from dev and test sets using \textit{\textbf{Seq2Seq$_\textbf{ASR}$}} that we train on the Librispeech ASR corpus.
Additionally, we use the $N$-best lists obtained from \cite{shin2019effective} in order to see the robustness of the language models on testing environments.
Table \ref{tab:wers} shows the word error rates (WERs) for each method after reranking.
The interpolation weights $\lambda$ were 0.3 or 0.4 in all $N$-best lists for ASR.

\begin{table}[t]
  \setlength\tabcolsep{3pt}
  \centering 
  \begin{tabular}{lcccc} 
    \toprule 
    \multicolumn{1}{c}{\multirow{2}{*}{Method}} & \multicolumn{2}{c}{dev} &  \multicolumn{2}{c}{test} \\ \cline{2-5} 
     &clean &other &clean &other \\ 
    
    \midrule
    \textit{\citeauthor{shin2019effective}} &7.17 &19.79 &7.25 &20.37 \\ 
    ~w/ n-gram &5.62 &16.85 &5.75 &17.72\\
    ~w/ $*${\small uniSANLM$_\text{w}$} &6.05 &17.32 &6.11 &18.13\\
    ~w/ $*${\small biSANLM$_\text{w}$} &5.52 &16.61 &5.65 &17.37\\
    ~w/ BERT &5.24 &16.56 &5.38 &17.46\\
    ~w/ BERT$_{\text{\textbackslash} M}$ &7.08 &19.61 &7.14 & 20.18\\
    \hdashline
    ~w/ uniLM\Tstrut &5.07\Tstrut &16.20\Tstrut &5.14\Tstrut &17.00\Tstrut \\ 
    ~w/ biLM &\textbf{4.94} &\textbf{16.09} &5.14 &\textbf{16.81}\\ 
    ~w/ T-TA &4.98 &\textbf{16.09} &\textbf{5.11} &16.91\\ 
    
    \midrule 
    \textit{\textbf{Seq2Seq$_\textbf{ASR}$}} &4.11 &12.31 &4.31 &13.14 \\ 
    ~w/ n-gram &3.94 &11.93 &4.15 &12.89\\
    ~w/ BERT &3.72 &11.59 &\textbf{3.97} &12.46\\
    ~w/ BERT$_{\text{\textbackslash} M}$ &4.09 &12.26 &4.28 & 13.15\\
    \hdashline
    ~w/ uniLM\Tstrut &3.82\Tstrut &11.73\Tstrut &4.05\Tstrut &12.63\Tstrut\\ 
    ~w/ biLM &3.73 &\textbf{11.53} &\textbf{3.97} &12.41\\ 
    ~w/ T-TA &\textbf{3.67} &11.56 &\textbf{3.97} &\textbf{12.38}\\ 
    
    \bottomrule
  \end{tabular}
  \caption{WERs after reranking with each language model on LibriSpeech. `other' sets are recorded in noisier environments than `clean' sets. Bolds are for the best performance on each sub-task. $*$ are word-level language models from \cite{shin2019effective}.}
  \label{tab:wers} 
\end{table}

We observe that the bidirectional language models trained with the LAE (T-TA) and MLM (biLM) outperform the unidirectional language model (uniLM) trained with the CLM.
Performance gains from the reranking are much lower in the better base system \textit{\textbf{Seq2Seq$_\textbf{ASR}$}}, and we can see that it is challenging to rerank the $N$-best list using a language model if the speech recognition model performs well enough.
Interestingly, the T-TA is competitive and even better than the biLM, and it may be from the gap between training and testing of the biLM: the biLM predicts multiple masks at a time when training, but predicts only one mask at a time when testing.
Moreover, the 3-layer T-TA is better than the 12-layer BERT-base, showing that in-domain data is critical to the language model applications.

Finally, we note that feeding an intact input to the BERT, denoted as ``w/ BERT$_\text{\textbackslash M}$'' in the Table \ref{tab:wers}, underperforms the others, and this shows that the \textit{mask-and-predict} is necessary for the effective reranking.

\subsubsection{Results on Machine Translation}

To see the reranking performance in other domain, NMT, we prepare the $N$-best lists using \textit{\textbf{Seq2Seq$_\textbf{NMT}$}}\footnote{Seq2Seq models for De$\rightarrow$En and Fr$\rightarrow$En are trained independently from t2t library \cite{vaswani2018tensor2tensor}} from the WMT-13's German-to-English and French-to-English test sets.
Table \ref{tab:bleus} shows the BLEU scores for each method after reranking.
Each interpolation weight becomes a value that shows the best performance on each test set with each method in NMT.
The interpolation weights $\lambda$ were 0.4 or 0.5 in the $N$-best lists for NMT.

\begin{table}[t]
  \centering
  \begin{tabular}{lcc} 
    \toprule 
    \multicolumn{1}{c}{Method} &De$\rightarrow$En &Fr$\rightarrow$En\\
    \midrule
    \textit{\textbf{Seq2Seq$_\textbf{NMT}$}} &27.83 &29.63\\
    ~w/ n-gram &28.41 &30.04\\
    ~w/ BERT &\textbf{29.31} &\textbf{30.52}\\
    \hdashline
    ~w/ uniLM\Tstrut &28.80\Tstrut &30.21\Tstrut \\
    ~w/ biLM &28.76 &\underline{30.32}\\
    ~w/ T-TA &\underline{28.83} &30.20\\
    \bottomrule
  \end{tabular}
  \caption{BLEU scores after reranking with each language model on WMT13. Bolds are for the best performance on each sub-task. Underlines are for the best in our implementations.}
  \label{tab:bleus}
\end{table}

We observe again that the bidirectional language models trained with the LAE and MLM perform better than the unidirectional language model trained with the CLM.
Also, the Fr$\rightarrow$En translation has less effect on reranking than the De$\rightarrow$En translation because the base NMT system for Fr$\rightarrow$En is better than that for De$\rightarrow$En.

Seeing that the $12$-layer BERT is much better than the others in reranking on NMT, it seems that the $N$-best hypotheses of the NMT model are more subtle to distinguish than those of the ASR model from the language model perspective.
All reranking results in ASR and NMT demonstrate that the proposed T-TA performs efficiently like uniLM and effectively like biLM.

\subsection{Unsupervised Semantic Textual Similarity}
\label{ssec:sts}
In addition to the reranking task, we apply language models to the semantic textual similarity (STS), which is the task of measuring the meaning similarity of sentence pairs.
We use STS Benchmark \cite{cer2017semeval} and SICK \cite{marelli2014semeval}, where both datasets have a set of sentence pairs with corresponding similarity scores.
The evaluation metric of STS is the Pearson's $r$ between the predicted similarity scores and the reference scores of the given sentence pairs.

In this section, we address the task of \textit{unsupervised} STS to examine the inherent ability to obtain contextual language representations of each language model, and we mainly compare language models that are trained on the English Wikipedia dump.
To compute a similarity score of a given sentence pair, we use the cosine similarity of two sentence representations, where each representation is obtained by averaging each language model's contextual language representations.
Specifically, contextual representations of a given sentence are the outputs of the final encoding layer of each model, denoted as \textit{context} in Table \ref{tab:sts} and \ref{tab:sick}.
For comparison, we use non-contextual representations, which are obtained from the outputs of the embedding layer, denoted as \textit{embed} in Table \ref{tab:sts} and \ref{tab:sick}.
As a strong baseline for unsupervised STS tasks, we also include the 12-layer BERT model \cite{devlin2019bert}, and we use the BERT in the \textit{mask-and-predict} approach for computing contextual representations of each sentence.
Note that we use the most straightforward approach for the unsupervised STS in order to focus on comparing token-level language representations.


\begin{table}[t]
  \setlength\tabcolsep{3pt}
  \centering
  \begin{tabular}{lcccc} 
    \toprule 
    \multicolumn{1}{c}{\multirow{2}{*}{Method}} & \multicolumn{2}{c}{STSb-dev} & \multicolumn{2}{c}{STSb-test}\\
    \cline{2-5}
    &\textit{context} &\textit{embed} &\textit{context} &\textit{embed}\\
    \midrule
    BERT &\textbf{64.78} &- &\textbf{54.22} &-\\
    BERT$_{\text{\textbackslash} M}$ &59.17 &60.07 &47.91 &48.19\\
    BERT$_{\texttt{[CLS]}}$ &\multicolumn{2}{c}{29.16} &\multicolumn{2}{c}{17.18}\\
    \hdashline
    uniLM\Tstrut &56.25\Tstrut &\textbf{63.87}\Tstrut &39.57\Tstrut &\textbf{55.00}\Tstrut \\
    uniLM$_{\texttt{[EOS]}}$ &\multicolumn{2}{c}{40.75} &\multicolumn{2}{c}{38.30}\\
    biLM &59.99 &- &50.76 &-\\
    biLM$_{\text{\textbackslash} M}$ &53.20 &58.80 &36.51 &49.08\\
    T-TA  &\textbf{71.88} &54.75 &\textbf{62.27} &44.74\\
    \hdashline
    GloVe\Tstrut &- &52.4\Tstrut &- &40.6\Tstrut \\
    Word2Vec &- &\textbf{70.0} &- &\textbf{56.5}\\
    \bottomrule
  \end{tabular}
  \caption{Pearson's $r\times100$ results on STS Benchmark. - denotes the infeasible value. Bolds are for the top-2 performances on each sub-task.}
  \label{tab:sts} 
\end{table}
\subsubsection{Results on STS Benchmark}
The STS Benchmark (STSb) has 5749/1500/1379 sentence pairs for train/dev/test splits with corresponding scores ranging from 0-5.
We test language models on the STSb-dev and STSb-test using the most simple approach on the unsupervised STS.
As our additional baselines, we include the results of GloVe \cite{pennington2014glove} and Word2Vec \cite{mikolov2013efficient} from the official sites of STS Benchmark\footnote{http://ixa2.si.ehu.es/stswiki/index.php/STSbenchmark}.

Table \ref{tab:sts} shows our T-TA trained with the LAE best captures the semantic of a sentence over the Transformer-based language models.
It is remarkable that our 3-layer T-TA trained on the relatively small data outperforms the 12-layer BERT trained on large data (Wikipedia + BookCorpus).
Another interesting point is that embedding representations are trained better by the CLM than the other language modeling objectives, and we guess that the uniLM highly depends on the embedding layer due to its constraint of the unidirectional context.

Since the uniLM encodes all contexts in the last token \texttt{[EOS]}, we also use the last representation as to the sentence representation, but it does not outperform the averaged sentence representation.
Similarly, BERT has a special token \texttt{[CLS]}, which is trained for the ``next sentence prediction'' objective, so we also use it to see how \texttt{[CLS]} learns sentence representation, but it significantly underperforms the others.


\begin{table}[t]
  \centering
  \begin{tabular}{lcc} 
    \toprule 
    \multicolumn{1}{c}{\multirow{2}{*}{Method}} & \multicolumn{2}{c}{SICK-test}\\
    \cline{2-3}
    &\textit{context} &\textit{embed}\\
    \midrule
    BERT  &64.31 &-\\
    BERT$_{\text{\textbackslash} M}$ &61.18 &64.63\\
    \hdashline
    uniLM\Tstrut  &54.20\Tstrut &\textbf{65.69}\Tstrut \\
    biLM  &58.98 &-\\
    biLM$_{\text{\textbackslash} M}$ &53.79 &62.67\\
    T-TA  &\textbf{69.49} &60.77\\
    \bottomrule
  \end{tabular}
  \caption{Pearson's $r\times100$ results on SICK data. - denotes the infeasible value. Bolds are for the best performance on each sub-task.}
  \label{tab:sick}
\end{table}

\subsubsection{Results on SICK}
We further evaluate language models on the SICK data, which consists of 4934/4906 sentence pairs for train/test splits with the scores ranging from 1-5.
The results are in Table \ref{tab:sick}, and we have the same observations as STSb.

All results on unsupervised STS tasks demonstrate that the T-TA learns textual semantics best using the token-level language modeling, LAE.

\section{Conclusion}

In this work, we propose a novel deep bidirectional language model named Transformer-based Text Autoencoder (T-TA) in order to eliminate the computational overload of applying BERT for unsupervised applications.
Experimental results on the $N$-best list reranking and the unsupervised semantic textual similarity tasks demonstrate that the proposed T-TA is significantly faster than the BERT-based approach, while its encoding ability is competitive or even better than that of BERT.

\section*{Acknowledgments}
K. Jung is with ASRI, Seoul National University, Korea. This work was supported by the Ministry of Trade, Industry \& Energy (MOTIE, Korea) under Industrial Technology Innovation Program (No.10073144) and by the NRF grant funded by the Korea government (MSIT) (NRF2016M3C4A7952587).

\bibliography{acl2020}
\bibliographystyle{acl_natbib}

\clearpage
\newpage

\appendix

\section{Appendices}
\subsection{Setups for ASR systems}
\label{a:ssec:asr}
This section introduces our implementation of the speech recognition system.

For the input features, we use 80-band Mel-scale spectrogram derived from the speech signal.
The target sequence is processed in 5K case-insensitive sub-word units created via unigram byte-pair encoding \cite{shibata1999byte}.
We use an attention-based encoder-decoder model as our acoustic model.
The encoder is a 5-layer bidirectional LSTM, and there are bottleneck layers that conduct linear transformation between every LSTM layers.
Also, there is a VGG module before the encoder, and it reduces encoding time steps by a quarter through two max-pooling layers.
The decoder is 2-layer bidirectional LSTM with location-aware attention mechanism \cite{chorowski2015attention}.
All the layers have 1024 hidden units.
The model is trained with additional CTC objective function because the left-to-right constraint of CTC helps learn alignments between speech-text pairs \cite{hori2017advances}.

Our model is trained for 20 epochs on 960h of LibriSpeech training data using Adadelta optimizer \cite{zeiler2012adadelta}.
Using this acoustic model, we obtain 50-best decoded sentences for each input audio through hybrid CTC-attention based scoring \cite{hori2017advances} method. For \textit{\textbf{Seq2Seq$_\textbf{ASR}$}}, we additionally use a pre-trained RNNLM to combine the log-probability $p^{lm}$ of RNNLM during decoding as follows:
\begin{align}
 & \log p(y_n|y_{1:n-1}) \nonumber \\
 & = \log p ^{\text{am}} (y_n|y_{1:n-1}) + \beta \log p^{\text{lm}}(y_n|y_{1:n-1}),
\end{align}
where $\beta$ is set to 0.7.
We use ESPNet toolkit \cite{watanabe2018espnet} for this implementation.

\begin{table}[th]
  \centering
  \begin{tabular}{ccccc}
    \toprule
    \multirow{2}{*}{Method} & \multicolumn{2}{c}{dev} & \multicolumn{2}{c}{test} \\ \cline{2-5}
    &clean\Tstrut  &other  &clean  &other \\
    \midrule
    \textit{\citeauthor{shin2019effective}} &7.17 &19.79 &7.26 &20.37 \\
    oracle &3.18 &12.98 &3.19 &13.61 \\
    \midrule
    \textit{\textbf{Seq2Seq$_\textbf{ASR}$}} &4.11 &12.31 &4.31 &13.14 \\
    oracle &1.80 &7.90 &1.96 &8.39 \\
    \bottomrule
  \end{tabular}
  \caption{Oracle WERs of the 50-best lists on LibriSpeech from each ASR system.}
  \label{a:tab:wer}
\end{table}

Table \ref{a:tab:wer} shows the oracle word error rates (WERs) of the 50-best lists, which are measured assuming that the best sentence is always picked from the candidates.
We also include the oracle WERs from the 50-best lists of \cite{shin2019effective}.

\subsection{Setups for NMT systems}
\label{a:ssec:nmt}

We implement the standard Transformer model \cite{vaswani2017attention} using Tensor2Tensor library \cite{vaswani2018tensor2tensor} for machine translation.
Both the encoder and decoder of the Transformer consist of 6 layers with 512 hidden units, and the number of the self-attention heads is 8.
The maximum number of input tokens is set to 256.
We use the shared vocabulary of size 32k.
For effective training, we let the token embedding layer and the last softmax layer share their weights.
The other hyperparameters of our translation system follow the standard \verb|transformer_base_single_gpu| setting in Google's official Tensor2Tensor repository\footnote{https://github.com/tensorflow/tensor2tensor}.

We train the baseline model on the standard WMT18 French-English and German-English datasets for 250k steps using Adam optimizer \cite{kingma2014adam}.
We use linear-warmup-square-root-decay learning rate scheduling with the default learning rate 2.5e-4 and warmup steps 16k.
Using this baseline translation model, we obtain 50-best decoded sentences for each source through the beam-search. 
The oracle BLEU scores for the NMT system are shown in Table \ref{a:tab:bleu}.

\begin{table}[th]
  \centering
  \begin{tabular}{ccccc}
    \toprule
    \multirow{2}{*}{Method} & \multicolumn{2}{c}{WMT13}\\ \cline{2-5}
     &De$\rightarrow$En\Tstrut &Fr$\rightarrow$En\\
    \midrule
    \textit{\textbf{Seq2Seq$_\textbf{NMT}$}} &27.83 &29.63 \\
    oracle &38.18 &39.58 \\
    \bottomrule
  \end{tabular}
  \caption{Oracle BLEUs of the 50-best lists on WMT}
  \label{a:tab:bleu}
\end{table}

\subsection{Running time of uniLM and T-TA}
\label{a:ssec:runtime}

\begin{figure}[t]
    \centering
    \includegraphics[width=\linewidth]{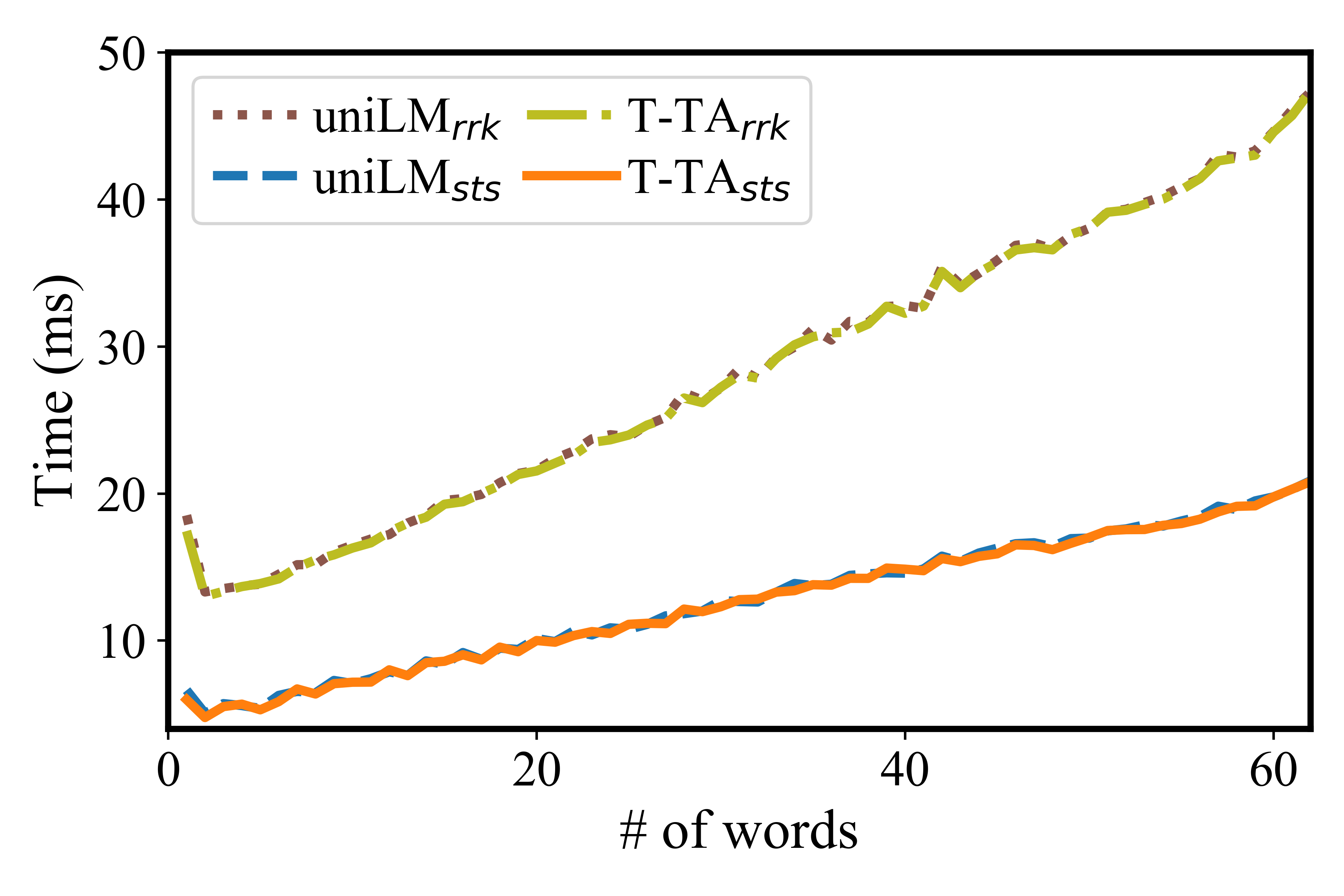}
    \caption{Running times according to the number of words for uniLM and T-TA.}
    \label{fig:times-unilm}
\end{figure}

As mentioned in Section \ref{ssec:runtime}, we also measure execution times of the uniLM we implement.
Figure \ref{fig:times-unilm} shows that the averaged run-times of the uniLM and the T-TA for the number of words in a sentence.
Since we use subword tokens, the number $n_w$ of words and the number $n$ of tokens can be different $n_w\leq n$.

\newpage

\subsection{Running time on GPU}
\label{a:ssec:runtime-gpu}

\begin{figure}[t]
    \centering
    \includegraphics[width=\linewidth]{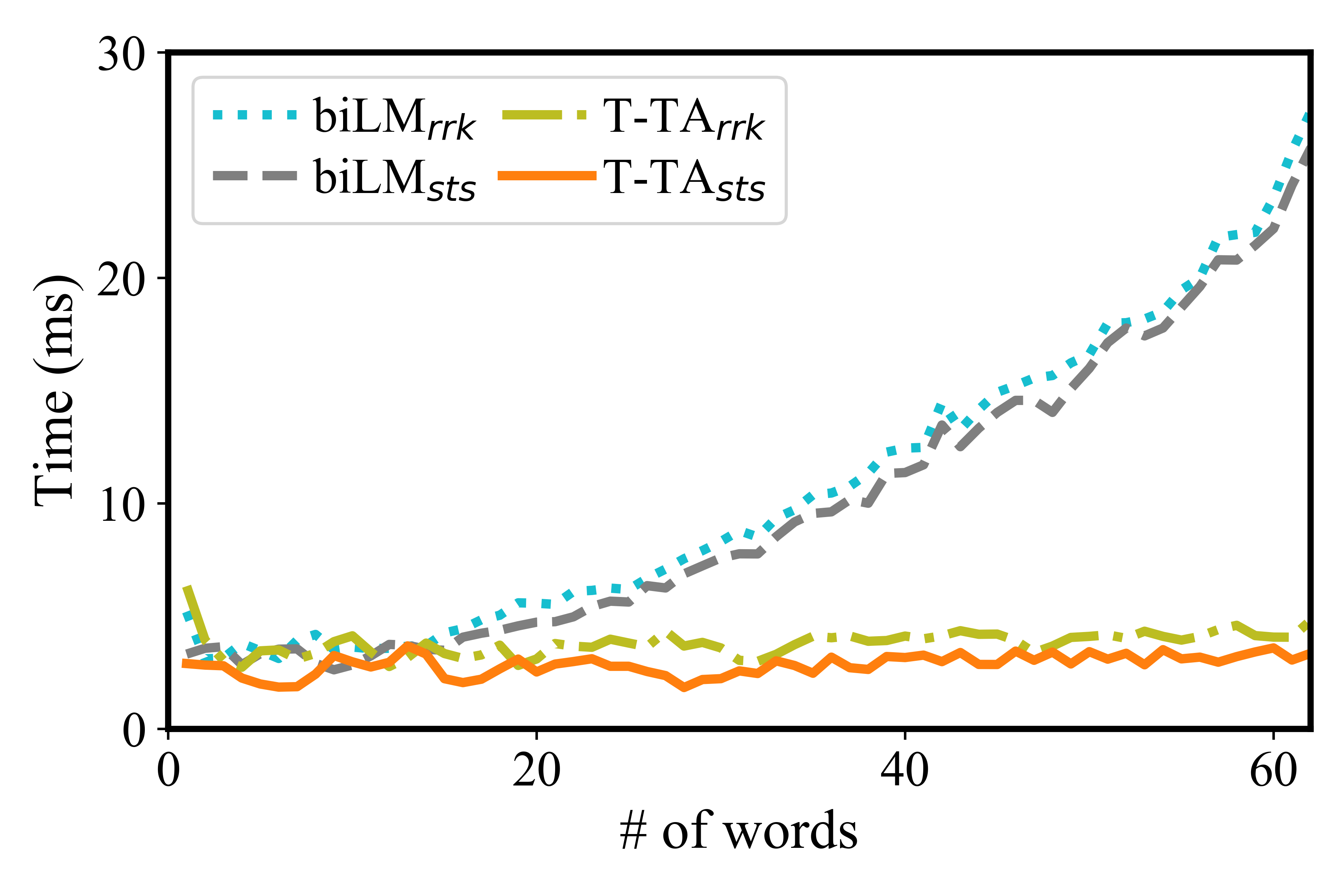}
    \caption{Running times according to the number of words for biLM and T-TA on GPU-augmented environment.}
    \label{fig:times-gpu}
\end{figure}

Additionally, we also measure execution times on a GPU-augmented environment (using GeForce GTX 1080 Ti).
Figure \ref{fig:times-gpu} shows that the averaged run-times of the biLM and the T-TA for the number of words in a sentence.
In our 20-words standard, the T-TA takes about $2.51$ ms and biLM takes about $4.72$ ms in the STS task, showing that the T-TA is $1.88$ times faster than the biLM.
Compared to the CPU-only environment, the speed difference was reduced due to the GPU supports.
Seeing Figure \ref{fig:times}, however, the CPU-only environment and the GPU-augmented environment have a similar tendency: the longer the sentence, the more significant the difference between the T-TA and the biLM.

\subsection{Perplexity and Reranking}
In general, perplexity (PPL) is a measure of how well language models trained.
To see the alignment of PPL and reranking, we compute PPL of reference sentences from the Librispeech dev-clean and test-clean set using each language model.
We can get pseudo-perplexity (pPPL) from biLM and T-TA since they do not follow the product rule, unlike uniLM.
We note that we compute subword-level (p)PPL (not word-level); these values are valid only in our vocabulary.

\begin{table}[th]
\setlength\tabcolsep{3pt}
  \centering 
  \begin{tabular}{crcc} 
    \toprule 
    \multicolumn{2}{r}{Method [{\small WER}]} &(p)PPL$_\text{a}$ &(p)PPL$_\text{m}$\\
    \midrule
    \multirow{3}{*}{\shortstack{dev\\clean}}
    &uniLM [3.82] &341.5   &70.80\\
    &biLM [3.73] &(76.49) &(11.93)\\
    &T-TA [3.67] &(293.4) &(11.69)\\
    \hline
    \multirow{3}{*}{\shortstack{test\\clean}}
    &uniLM [4.05] &495.5   &73.18\\
    &biLM [3.97] &(75.43) &(12.72)\\
    &T-TA [3.97] &(590.0) &(12.43)\\
    \bottomrule
  \end{tabular}
  \caption{(pseudo)Perplexities and corresponding WERs of language models on LibriSpeech.}
  \label{tab:ppls} 
\end{table}
We can find that WERs are better aligned with the median of pPPL$_\text{m}$ than the averaged pPPL$_\text{a}$.
Interestingly, the pPPL$_\text{a}$ of T-TA is similar to the PPL$_\text{a}$ of uniLM, but the pPPL$_\text{m}$ of T-TA is similar to that of biLM.
We additionally find that if the length of a sentence is short, T-TA shows a very high perplexity, even higher than uniLM.

\end{document}